\documentclass[conference]{IEEEtran}
\IEEEoverridecommandlockouts
\usepackage{cite}
\usepackage{amsmath,amssymb,amsfonts}
\usepackage{graphicx}
\usepackage{xcolor}
\usepackage{comment, lipsum}
\usepackage{amsmath,amssymb} 
\usepackage{tikz}
\usepackage{color}
\usepackage{multirow, multicol}
\usepackage{graphicx, caption, subcaption}
\usepackage{newtxtext,newtxmath}
\usepackage{subcaption}
\usepackage{newtxmath}
\usepackage{algorithm}
\usepackage{algpseudocode}
\usepackage{varwidth}
\usepackage{tabulary,booktabs}

\begin{document}

\title{WeightMom: Learning Sparse Networks using Iterative Momentum-based pruning
}

\author{
\IEEEauthorblockN{Elvis Johnson}
\IEEEauthorblockA{
\textit{University of Washington}\\
Washington, USA}
\and
\IEEEauthorblockN{Xiaochen Tang}
\IEEEauthorblockA{
\textit{New York University} \\
New York, USA}
\and
\IEEEauthorblockN{Sriramacharyulu Samudrala}
\IEEEauthorblockA{
\textit{Unipart Group} \\
Mumbai, India}

}

\maketitle

\begin{abstract}
Deep Neural Networks have been used in a wide variety of applications with significant success. However, their highly complex nature owing to comprising millions of parameters has lead to problems during deployment in pipelines with low latency requirements. As a result, it is more desirable to obtain lightweight neural networks which have the same performance during inference time. In this work, we propose a weight based pruning approach in which the weights are pruned gradually based on their momentum of the previous iterations. Each layer of the neural network is assigned an importance value based on their relative sparsity, followed by the magnitude of the weight in the previous iterations. We evaluate our approach on networks such as AlexNet, VGG16 and ResNet50 with image classification datasets such as CIFAR-10 and CIFAR-100. We found that the results outperformed the previous approaches with respect to accuracy and compression ratio. Our method is able to obtain a compression of 15$\times$ for the same degradation in accuracy on both the datasets.
\end{abstract}


\section{Introduction}
Deep Neural Networks (DNNs) have been highly useful in enabling significant advancements in a wide array of research areas ranging from applications like object recognition to music generation and biomedical applications \cite{balakrishnan2013detecting, li2014remote, subramaniam2019spectral}. As the popularity of deep neural networks increased, there has been a trend to apply deep learning to applications with resource constraints as well. For instance, it has been commonplace to expect optimal performance from a memory constrained device such as smartphones. As a result, it has become necessary to develop techniques to simplify the complex structure of DNNs while maintaining their test performance.

Initial efforts to reduce the total number of parameters in deep neural networks primarily involved eliminating the weights of the network based on their magnitude \cite{lecun1989optimal, reed1993pruning}. If the magnitude of the weight was lesser than a pre-specified threshold, then the weight was eliminated since it was assumed to be of minimal importance to the network. This was due to the fact that the weights would minimize the impact of its coefficients during matrix multiplications required for backward and forward propagation.
Over the years, developments in hardware has enabled the implementation of complex neural networks on specifically designed hardware accelerators \cite{deng2020model}. Furthermore, while certain papers have focused on improving the architecture required to run deep learning algorithms, others have focused on introducing new elements to the circuit which can take advantage of the general learning capability of a neuron. For instance, \cite{bojnordi2016memristive, subramaniam2017neuromorphic, boppidi2020implementation} make use of memristors to emulate the learning capability of a synapse. \cite{raj2019programming} uses different modulation techniques to program the learning capability of the memristor. While these techniques (hardware-based techniques) can be quite effective in improving the efficiency of deep neural networks while maintaining the inference accuracy, it can often take a long time to design and fabricate the hardware required for real world applications. Additionally, in case of an increase in the overall scale of the system, it would be impractical to wait for another hardware design. In such cases, it would be desirable to develop software-based approaches in which the model is pruned for the particular task, irrespective of the hardware on which it is implemented.\\
Recently, with the rising popularity of one shot pruning approaches (motivated by the Lottery Ticket Hypothesis \cite{frankle2018lottery}), the model is pruned in a single shot manner before training. There have been other approaches which have focused on one shot pruning.\cite{lee2018snip} trains the reference model on a single mini-batch before pruning the weights in a single shot. The pruned model is then trained on the current dataset to achieve convergence. 
The advantages of single shot pruning is that the pruned model is the pruned model more portable for multiple tasks and allows for a much simpler training procedure as compared to iterative pruning strategies, where the model is pruned and train simultaneously. Our work is a step in this direction.

In this work, we propose to apply momentum-based weight pruning in which we selectively prune the weights in the network based on their momentum in the last few epochs. As a result, we only prune the weights which have been consistently lower over a period of iterations, rather than pruning them based on a single iteration. We evaluate our method for models such as VGG16 and ResNet50, on CIFAR10 and CIFAR100. The proposed method outperforms previous state-of-the-art methods on both of the above models across multiple datasets. The details of our approach and the experimental results are illustrated in the following sections. \\
The rest of this paper is organized as follows. In Section II, we report the current related literature on deep neural network compression. Section III provides a formulation of our method. Section 4 reports the experimental results of our approach on VGG16 and ResNet50. Finally, Section IV concludes the paper.
\newpage

\section{Related Work}

The highly complicated structure of deep neural networks can often result in latency and memory issues during deployment in large-scale pipelines which deal with huge amounts of data on a daily basis. Furthermore, even for systems which have limited data, but need to be deployed on devices with resource constraints - deploying deep learning algorithms can be a significant challenge \cite{icaart}. 
Research efforts to mitigate this issue have primarily followed one of the two approaches to tackle the computation intensive nature of deep neural networks - (1) Hardware based methods and (2) Software-based Methods.
\subsection{Hardware based methods}
Several methods have attempted to target the hardware of the device which runs Deep Neural Networks and other computation intensive algorithms \cite{schuman2017survey, marculescu2018hardware, mittal2019survey}. Most of the previous papers have primarily focused on optimizing the hardware specifically for running Deep Neural Networks and other deep learning-based applications  \cite{ambrogio2018equivalent, deng2020model, subramaniam2017neuromorphic}, others have emphasized on a lower-level approach in which the essential components of the hardware circuit have been modified to obtain superior performance. This approach is akin to a \textit{ground-up rebuild} of sorts. For instance, \cite{boppidi2020implementation, bojnordi2016memristive, raj2019programming} make use of resistive RAM technology and components such as memristors to mimic the learning capability of a synapse. In other words, through a variety of signal processing techniques, it has been shown that it is possible to emulate properties such as Spike Timing Dependant Plasticity (STDP), thereby creating a circuit which can act as an artificial synapse. While these approaches can be helpful in bringing about a huge change in the overall performance of deep learning specialized hardware, the wait time for a successful implementation can be long since the time taken to fabricate and deploy such components can be unreliable and lead to delays in the downstream processes.  Additionally, it can become difficult to develop solutions which can dynamically adapt to the ever-changing scalability of the systems deployed in large-scale environments.  

\subsection{Software-based Methods}
Over the last few years, the rise of Deep Learning has lead to the need to come up with pruning approaches which can maintain the overall accuracy of the pruned model while reducing its time or memory footprint. This has resulted in a variety of approaches such as:

\begin{itemize}
    \item \textbf{Weight-based pruning} - In this approach, papers have iteratively eliminated weights with magnitude less than a specified threshold under the assumption that weights with lower magnitude do not significantly affect the overall test performance of the model  \cite{lecun1989optimal, hassibi1992second}.  Weight-based pruning can further be classified into different kinds of approaches based on the relative ordering of the training and the pruning procedures: \\
    \begin{enumerate}
        \item \textbf{Pruning first}: Recently, inspired by the Lottery Ticket Hypothesis \cite{frankle2018lottery}, a higher number of papers have focused on pruning before training. For instance, \cite{lee2018snip} focuses on training the model in one shot before training the model. In order to obtain some prior information before pruning the model, they train the model on a single mini-batch of data before pruning the weights. \cite{subramaniam2020n2nskip} is a novel approach in which sparsely connected convolutional skip connections are employed increase the inherent connectivity of the model.  
        \item \textbf{Training first}: The two most popular approaches to training-first followed by pruning are Hessian pruning and weight-magnitude pruning. These have mainly been spurred by \cite{hassibi1992second} and \cite{lecun1989optimal}. In this approach, the model is first trained until it achieves convergence on the task (such as image classification), following which it is pruned. Next, the pruned model is fine-tuned on the task until it achieves the previous (or slightly lower) accuracy.
        \item \textbf{Prune-Train-Prune}: This has been the most popular pruning approach to date \cite{mallya2018packnet, dettmers2019sparse, liu2017learning, peng2019collaborative, dong2017learning}. In this approach, each reference model is trained for a few epochs followed by a pruning step every $n$ epochs. As a result, the sparsity of the model is gradually increased, due to which the accuracy of the model does not drop drastically. Additionally,
        \cite{narang2017exploring} proposed a schedule to optimally pruned the the model depending on the number of layers, parameters and the ratio of convolutional to fully connected parameters in the model.
    \end{enumerate}
    \item \textbf{Channel-based Pruning} - In this method, the neurons/channels of the network are pruned based on the overall set of weights or filters each channels receives. One such prominent approach in channel pruning uses the concept of network slimming \cite{liu2017learning}. Each channel is pruned based on the scaling factor in the batch-norm associated with the corresponding channel. As a result, this leads to a \textit{slimming} of the network, where the number of channels in each layer of the network reduces or becomes zero. A drawback of this approach is that it may be difficult to control the overall sparsity of the model (as compared to weight-based pruning) since removing a single channel can remove a large number of connections based on the layer in which it is present.
    \item \textbf{Filter-based Pruning} - One such method which has become popular over the years has been filter-based pruning. In this approach, each individual coefficient of the filters in a convolutional neural network are removed, thereby reducing the overall computation arising from a large number of matrix multiplications.
\end{itemize}

\noindent
Recently, \cite{dettmers2019sparse, ding2019global} have employed momentum in the gradients of the model to rank each model parameter based on the importance of its corresponding derivative. In other words, the weight which consistently had a higher gradient update was assigned a higher importance and thereby, preserved in the pruned model.

\section{Method}

We apply iterative weight based pruning based on the momentum of the weight magnitude over the last 15 epochs. As a result, the weights removed are the ones whose magnitudes have been consistently lower that a pre-specified threshold for over 10-15 epochs. \\
Additionally, we realize that each layer of a deep neural network is not the same. For instance, the initial layers would be more important to the learning of the neural network since they capture low-level features such as shapes and colors, while the final layers capture more advanced features. Hence, we formulate an importance ratio for each layer based on their relative position in the network and the total number of parameters in each layer as compared to the total number of parameters in the deep neural network. The importance ratio is given by:

\begin{equation}
    I(W, l) = \frac{W_{l}}{l\cdot W_{avg}}
\end{equation}

where $l$ is the position of the layer from the beginning, $W_{l}$ is the number of parameters in layer $l$ and $W_{avg}$ is the average number of parameters in a layer in the deep neural network. In other words, $W_{avg}$ can be expressed as:
\begin{equation}
    W_{avg} = \frac{W_{l}}{\Sigma_{i} W_{i}}
\end{equation}

We would like to note that $W_{avg}$ is an important parameter that determines the relative importance of the layer with respect to other layers since layers with a higher number of parameters can allow for a higher compression ratio, thereby enabling a higher compression rate while maintaining the overall test accuracy.

\section{Experimental Results}
\subsection{Experimental Setup}
\subsubsection{Datasets}
The proposed method is evaluated on the following three datasets:
\begin{itemize}
    \item \textbf{CIFAR-10} - The CIFAR-10 datasets consists of 50,000 images as part of the training data and 10000 images as part of the test data. It has a total of 10 classes.
    \item \textbf{CIFAR-100} - The CIFAR-100 datasets consists of 50,000 images as part of the training data and 10000 images as part of the test data. It has a total of 100 classes.
\end{itemize}

For evaluating our approach, we have considered CIFAR-10 and CIFAR-100 since these datasets have been evaluated on by a majority of previous methods. As a result, we would be able to compare the robustness of our approach directly against previous state-of-the-art (SOTA) approaches. 

\subsubsection{Models}
Similar to the setup followed in previous approaches, we focus on popular architectures, namely AlexNet, VGG16 and ResNet50. Each of these models is first trained without any pruning to obtain the initial baseline accuracy of the reference model. We have chosen 2 other approaches as competitive baselines in order to evaluate the efficacy of our proposed approach.

\subsubsection{Training Schedules}
Each model is trained with the following hyperparameters:
We use an initial learning rate of 0.05, with a learning rate decay of 0.5. We decay the learning rate every 30 epochs so that the model does not get stuck at a local optima. We apply Adam optimizer with a momentum of 0.9. Finally, we use a batch size of 128 so as to be in sync with previous papers. The results reported in Tables \ref{tab:cifar10} and \ref{tab:cifar100} are based on three runs of each model at each compression ratio in order to ensure that our results are statistically significant. Each model is trained on four NVIDIA
1080Ti GPUs  before pruning so that we have the baseline accuracies of each model in the respective dataset. Next, we apply pruning in the model and check the degradation in accuracy at three difference compression ratios - 10$\times$, 20$\times$ and 50$\times$. 

\subsection{CIFAR-10}
Table ~\ref{tab:cifar10} reports the experimental results of the proposed method on AlexNet, ResNet and VGG on the CIFAR-10 dataset. We document the degradation in accuracy at three different compression ratios. These are 10$\times$, 20$\times$ and 50$\times$. These compression ratios correspond to a network density where we retain 10\%, 5\% and 2\% of the overall parameters. As we can see from Table ~\ref{tab:cifar10}, the proposed method outperforms the previous approaches at all three network densities. Furthermore, our method obtains the minimal decrease in accuracy at extremely high sparsities of 5\% and 2\%.

\setlength{\tabcolsep}{3.5pt}
\begin{table}[htbp]
\begin{center}
\begin{tabular}{ccccc}
\hline\noalign{\smallskip}
\multirow{2}{*}{Model} & \multirow{2}{*}{Method} & & Density &  \\ 
\cline{3-5}
& & 10\% & 5\% & 2\% \\
\hline
\vspace{1mm}
\multirow{5}{*}{AlexNet} & {Baseline} & $83.50$ & - & -\\
\vspace{1mm}
 & {SNIP \cite{lee2018snip}} & $73.42 $ &  $70.42$ & $68.67 $\\
 \vspace{1mm}
 & {N2NSkip \cite{subramaniam2020n2nskip}} & $73.42$ &  $70.42$ & $68.67$\\
 \vspace{1mm}
 & {\textbf{Ours}} & $\boldsymbol{74.59}$ & $\boldsymbol{72.89}$ & $\boldsymbol{72.09}$\\
\hline
\vspace{1mm}
\multirow{5}{*}{VGG16} & {Baseline} & $83.50$ & - & -\\
\vspace{1mm}
 & {SNIP \cite{lee2018snip}} & $73.42 $ &  $70.42$ & $68.67 $\\
 \vspace{1mm}
 & {N2NSkip \cite{subramaniam2020n2nskip}} & $73.42$ &  $70.42$ & $68.67$\\
 \vspace{1mm}
 & {\textbf{Ours}} & $\boldsymbol{74.59}$ & $\boldsymbol{72.89}$ & $\boldsymbol{72.09}$\\
\hline
\vspace{1mm}
\multirow{5}{*}{ResNet50} & {Baseline} & $83.50$ & - & -\\
\vspace{1mm}
 & {SNIP \cite{lee2018snip}} & $73.42 $ &  $70.42$ & $68.67 $\\
 \vspace{1mm}
 & {N2NSkip \cite{subramaniam2020n2nskip}} & $73.42$ &  $70.42$ & $68.67$\\
 \vspace{1mm}
 & {\textbf{Ours}} & $\boldsymbol{74.59}$ & $\boldsymbol{72.89}$ & $\boldsymbol{72.09}$\\
\hline
\end{tabular}
\end{center}
\caption{Test Accuracy of pruned AlexNet, VGG16, ResNet50 on CIFAR-10.}
\label{tab:cifar10}
\end{table}

\subsection{CIFAR-100}

\setlength{\tabcolsep}{3.5pt}
\begin{table}[htbp]
\begin{center}
\begin{tabular}{ccccc}
\hline\noalign{\smallskip}
\multirow{2}{*}{Model} & \multirow{2}{*}{Method} & & Density &  \\ 
\cline{3-5}
& & 10\% & 5\% & 2\% \\
\hline
\vspace{1mm}
\multirow{5}{*}{AlexNet} & {Baseline} & $83.50$ & - & -\\
\vspace{1mm}
 & {SNIP \cite{lee2018snip}} & $73.42 $ &  $70.42$ & $68.67 $\\
 \vspace{1mm}
 & {N2NSkip \cite{subramaniam2020n2nskip}} & $73.42$ &  $70.42$ & $68.67$\\
 \vspace{1mm}
 & {\textbf{Ours}} & $\boldsymbol{74.59}$ & $\boldsymbol{72.89}$ & $\boldsymbol{72.09}$\\
\hline
\vspace{1mm}
\multirow{5}{*}{VGG16} & {Baseline} & $83.50$ & - & -\\
\vspace{1mm}
 & {SNIP \cite{lee2018snip}} & $73.42 $ &  $70.42$ & $68.67 $\\
 \vspace{1mm}
 & {N2NSkip \cite{subramaniam2020n2nskip}} & $73.42$ &  $70.42$ & $68.67$\\
 \vspace{1mm}
 & {\textbf{Ours}} & $\boldsymbol{74.59}$ & $\boldsymbol{72.89}$ & $\boldsymbol{72.09}$\\
\hline
\vspace{1mm}
\multirow{5}{*}{ResNet50} & {Baseline} & $83.50$ & - & -\\
\vspace{1mm}
 & {SNIP \cite{lee2018snip}} & $73.42 $ &  $70.42$ & $68.67 $\\
 \vspace{1mm}
 & {N2NSkip \cite{subramaniam2020n2nskip}} & $73.42$ &  $70.42$ & $68.67$\\
 \vspace{1mm}
 & {\textbf{Ours}} & $\boldsymbol{74.59}$ & $\boldsymbol{72.89}$ & $\boldsymbol{72.09}$\\
\hline
\end{tabular}
\end{center}
\caption{Test Accuracy of pruned AlexNet, VGG16, ResNet50 on CIFAR-100.}
\label{tab:cifar100}
\end{table}

Table ~\ref{tab:cifar100} reports the experimental results of the proposed method on AlexNet, ResNet and VGG on the CIFAR-100 dataset. We document the degradation in accuracy at three different compression ratios. These are 10$\times$, 20$\times$ and 50$\times$. These compression ratios correspond to a network density where we retain 10\%, 5\% and 2\% of the overall parameters. Similar to the performance on CIFAR-10, we observe that the proposed method outperforms the previous approaches at all three network densities. Additionally, our method obtains the minimal decrease in accuracy at extremely high sparsities of 5\% and 2\%.

\begin{figure}
    \centering
    \includegraphics[scale=0.5]{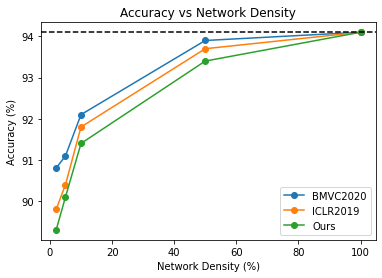}
    \caption{VGG16 - Degradation in test accuracy for VGG16 across BMVC2020, ICLR2019 and the proposed method with increasing network sparsity.}
    \label{fig:vggcifar10}
\end{figure}

\begin{figure}
    \centering
    \includegraphics[scale=0.5]{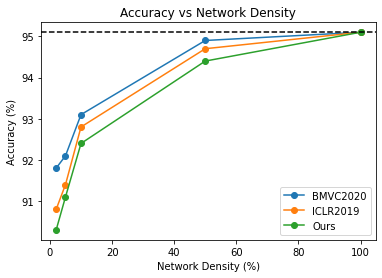}
    \caption{ResNet50 - Degradation in test accuracy for VGG16 across BMVC2020, ICLR2019 and the proposed method with increasing network sparsity.}
    \label{fig:resnetcifar10}
\end{figure}

There are a few observations that can be drawn from results in Tables \ref{tab:cifar10} and \ref{tab:cifar100}. 
\begin{itemize}
    \item \textbf{Our approach scales better at higher sparsities} - As shown in \ref{fig:resnetcifar10} and \ref{fig:vggcifar10}, our approach is able to retain a larger proportion of important connections as compared to BMVC2020 and ICLR2019, due to which it is more robust at higher compression rates.
    \item Our approach is more adept preserving overall connectivity in smaller and well as larger networks, while previous approaches are mostly effective only for larger networks.
\end{itemize}

\section{Conclusion}
We proposed WeightMom - an iterative weight-based pruning strategy based on the momentum of the weight magnitudes across the previous few iterations. We found that while maintaining the same test accuracy, our proposed approach was able to better preserve the important parameters in the network especially at extremely high compression rates of over 10\%.  We believe that using momentum to prune weights can be an extremely useful tool to determine which parameters are important to the network, since it takes the performance of the weight over a period of time before determining the importance of the parameter, rather than discarding the parameter based on a single examination.

\bibliography{pruning}


\end{document}